%% file: DAS_2016_script.tex
\documentclass[conference]{IEEEtran}
\pdfoutput=1
\usepackage{mwe}
\usepackage{graphbox} 
\usepackage{url}
\usepackage{amsmath}
\usepackage{amssymb}
\usepackage{color}
\usepackage{epstopdf}
\usepackage{tabularx}
\graphicspath{{./}{./images/}}
\usepackage{hyperref}
\usepackage{subfigure}
\usepackage{gensymb}
\usepackage{lipsum}
\usepackage[export]{adjustbox}
\usepackage{caption}
\usepackage{flushend}

\begin{document}

\title{Visual Script and Language Identification}

\author{\IEEEauthorblockN{
Anguelos Nicolaou\IEEEauthorrefmark{1},
Andrew D. Bagdanov\IEEEauthorrefmark{2},
Lluis Gomez-Bigorda\IEEEauthorrefmark{1}, and
Dimosthenis Karatzas\IEEEauthorrefmark{1}
}
\IEEEauthorblockA{
\IEEEauthorrefmark{1}
Computer Vision Center, Edifici O, Universitad Autonoma de Barcelona,Bellaterra, Spain
\\ 
\IEEEauthorrefmark{2}
Media Integration and Communication Center, University of Florence, Florence, Italy 
\\
Email: \href{mailto:anguelos@cvc.uab.es}{anguelos@cvc.uab.es},
\href{mailto:bagdanov@cvc.uab.es}{andrew.bagdanov@unifi.it}, 
\href{mailto:lgomez@cvc.uab.es}{lgomez@cvc.uab.es},
\href{mailto:dimos@cvc.uab.es}{dimos@cvc.uab.es}
}

}

\maketitle

\begin{abstract}
In this paper we introduce a script identification method based on hand-crafted texture features and an artificial neural network.
The proposed pipeline achieves near state-of-the-art performance for script identification of video-text and state-of-the-art performance on visual language identification of handwritten text.
More than using the deep network as a classifier, the use of its intermediary activations as a learned metric demonstrates remarkable results and allows the use of discriminative models on unknown classes.
Comparative experiments in video-text and text in the wild datasets provide insights on the internals of the proposed deep network.
\end{abstract}

\IEEEpeerreviewmaketitle
\section{Introduction}
\label{s:introduction}
As document analysis systems are evolving, their multi-lingual capabilities are becoming more important.
Script identification is a key element in multilingual system pipelines.
Other than performance in detecting the script and the language of text in such pipelines, the position this step occupies in the pipeline dictates whether it will assist or be assisted by other steps in the pipeline.


In this paper we address the problem of script or language identification in several modalities such as video-text, scene-text, or handwritten text,  and introduce a method consisting of hand-crafted features and a fully connected deep neural network. 
We demonstrate that k-NN classification over the features obtained from the first layer of the deep neural network equals or outperforms the deep network classification.
The principal contributions of this paper are: 
the introduction of a method that uses a deep neural network on top of hand-crafted features for script identification, a method to perform a purely visual identification of language, even for languages sharing the same script, and the use of the activations of the employed neural network as a learned metric in order to generate more adaptable classifiers.

\section{Background}
\label{s:background}

\subsection{Script Identification}
\label{s:script}
Script detection has been an open problem for several decades.
For the contents of this paper, script identification refers to identifying the system of writing, the alphabet used in a sample, while language identification refers to identifying the language given a text sample.
The above definition produces ambiguities on some cases, yet those two notions from a pattern recognition perspective are very different.
Script identification implies focussing on detecting symbols, while language identification implies detecting some specific auxiliary symbols, such as diacritics, and an underlying language model.
Several variations of the problem exist depending on aspects such as the granularity of the data samples, the number of scripts out of which the systems classify, and the modality of the textual data, i.e. whether its printed text, handwritten text, scene text etc.
For a detailed overview of script identification before 2009, we refer to \cite{ghosh2010script}, which provides a thorough taxonomy of methods available up to that time.
In 2009 Unnikrishnan and Smith~\cite{unnikrishnan2009combined} demonstrated that for simple cases of binarized printed text, the problem can be considered solved by a method developed for the Tessaract OCR engine.
Zhu et al.~\cite{zhu2009language} have used codebooks generated from printed and handwritten data in order to perform handwritten language identification.
Ferrer et al.~\cite{ferrer2013lbp} used the simple $LBP_{3 \times 3}$ pooled horizontally to perform script identification.
More recently Long Short Term Memory (LSTM) networks have been used by Ulhasan et al.~\cite{ulhasan2015sequence} for separating characters in multilingual text with a granularity of characters.
Mioulet et al \cite{mioulet2015language} also used a bidirectional variant of LSTM networks with a cascade of script detection, OCR and language models, in order to infer the language even when two languages share the same script.
The ICDAR2015 Competition on Video Script Identification (CVSI)~\cite{cvsi2015} posed the problem of script identification over superimposed text in videos
; the four best participant methods were all using Convolutional Neural Networks (CNN).
Shi et al.~\cite{shi2015script} have also used a deep CNN to address the problem of script detection in the wild.
The CNN approach has the drawback of the need for vast computational resources as well as large amounts of annotated data.
Depending on the granularity of samples, i.e. character based, word based, text-line based, and paragraph based, as well as the modality of text, whether it be scene-text, printed documents, or handwritten texts, script identification can be seen as many problems rather than one.
From this perspective, the problem addressed in this paper, script identification of a word level granularity on scene-text, is a challenging one that is starting to gain momentum.

\subsection{Text as texture}
\label{s:text}
This paper builds on previous work that introduced the Sparse Radial Sampling (SRS) variant of the Local Binary Pattern (LBP)~\cite{nicolaou2015sparse}.
Classifying text regions by global texture descriptors is a strategy that proved effective in the task of writer identification and now is employed and used in a different problem and in the context of supervised learning.
The strategy of analysing text using texture analysis, apart from yielding good results has the benefit of providing this information at an early stage in a pipeline and making it available to following stages of a text identification.

\section{Method}
\label{s:method}
The proposed method consists of a preprocessing step, followed by LBP feature extraction, and training an Artificial Neural Network (ANN) on these features.
The intermediary layers of the ANN are then used as a generative model to perform classification.

\subsection{Preprocessing}
\label{s:preprocessing}
\input{fig_features}
Before passing images to the LBP transform, each image is preprocessed independently.
Since the LBP transform is applied on a single channel image, instead of luminance, the principal component of all pixel colors was chosen in order to enhance the perceptual differences that are not attributed to luminance.
In order to have a consistent LBP encoding between images with a light foreground on a dark background and images with a dark foreground on a light background, whenever the central band is darker than the image average, the image is flipped.
The assumption is that more foreground pixels will exist in the central band between 25\% and 75\% of the image width.
In fig.~\ref{fig:features} rows 1) and 2) demonstrate some examples of the preprocessing.
When global pooling and using local structure features such as the LBP or Histogram of Oriented Gradients (HOG), inverting the image has an effect equivalent to flipping it across both axes.
This means that making all samples have light background on dark foreground is as important as enforcing all samples to be properly oriented. 
In Column (b) of Fig.~\ref{fig:features} the effect the flipping has on the LBP transform can be seen.

\subsection{Local Binary Patterns}
\label{ss:local}
For feature extraction the SRS-LBP variant of LBP histogram features is employed.
Briefly, the SRS-LBP embeds a clustering of the center-neighbourhood differences using Otsu's~\cite{otsu1975threshold} method; it also uses a dis-joined approach to obtain a multi-radius feature representation with linear complexity.
LBP have several advantages for script identification: they exploit the bi-level nature textual images have, they are very fast to compute, and they are pooled over regions which makes them segmentation-free and an inherently global descriptor of an image region.
In~\cite{ferrer2013lbp} Ferrer et al. extracted LBP features from text-lines by concatenating histograms of 4 horizontal stripes.
In~\cite{shi2015automatic} Shi et al. used a deep convolutional network that employs horizontal pooling to discard spatial information along the horizontal direction.

In the same respect, and assuming images are either cropped words or cropped lines, the SRS-LBP were extracted for 3 regions in the images: the upper half of the image, the central half of the image, and the lower half of the image.
The dimensionality of the extracted features-set is the product of the histogram size, the different radii and the pooling zones: $2^8 \times 12 \times 3 = 9216$.
In Fig.~\ref{fig:features}, in column (d) a rare example where the SRS-LBP is fooled can be seen; this happens because the SRS-LBP assumes that the most significant contrast in an image with text will be the contrast related to foreground-background transitions.

\subsection{Classification}
\label{s:classification}
\begin{figure}
\includegraphics[width=\columnwidth]{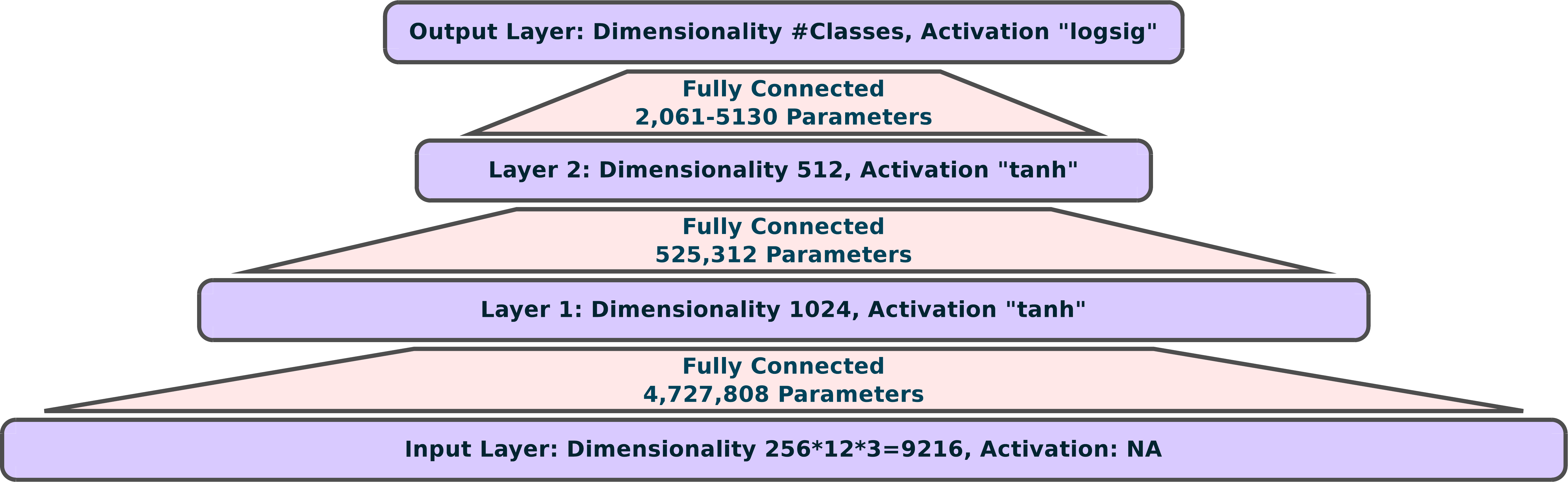}
\caption{Architecture of the Proposed Neural Network}
\label{fig:architecture}
\end{figure}

The remaining pipeline of the SRS-LBP is an unsupervised learning approach, aimed at totally different class and samples per class cardinalities.
A deep Multi Layer Perceptron (MLP) is used as a classifier of the feature representation to a given and limited set of languages.
The network consists of 3 fully connected layers plus the input layer.
The first layer maps the 9,216 features to a dimensionality of 1,024, the second layer maps the data from 1,024 to 512 dimensions and the third layer maps the data to as many neurons as the number of classes.
The output layer can be interpreted as the probability of the presented data belonging to each class.
The activations for the layers are respectively $tanh$, $tanh$, and the logistic function.
In Fig.~\ref{fig:architecture} a visual representation of the architecture can be seen.
It should be pointed out that the number of parameters of the network varies depending on the feature vectors dimensionality as well as the number of classes in each dataset used.
In the case of classifying word images to 10 classes, the model has 9,968,138 parameters.

For training\footnote{For all experiments, the KERAS~\cite{keras} framework was used.}, Stochastic Gradient Descent (SGD) is used with \textit{categorical cross-entropy} as a loss function.
Drop-out regularisers of 0.5 are used on each layer~\cite{srivastava2014dropout}.
The batch-size is set to be proportional to the the number of samples per class, but no less than 32.

\subsection{MLP as Metric Learning}
\label{s:knn}
\begin{figure}
\includegraphics[width=\columnwidth]{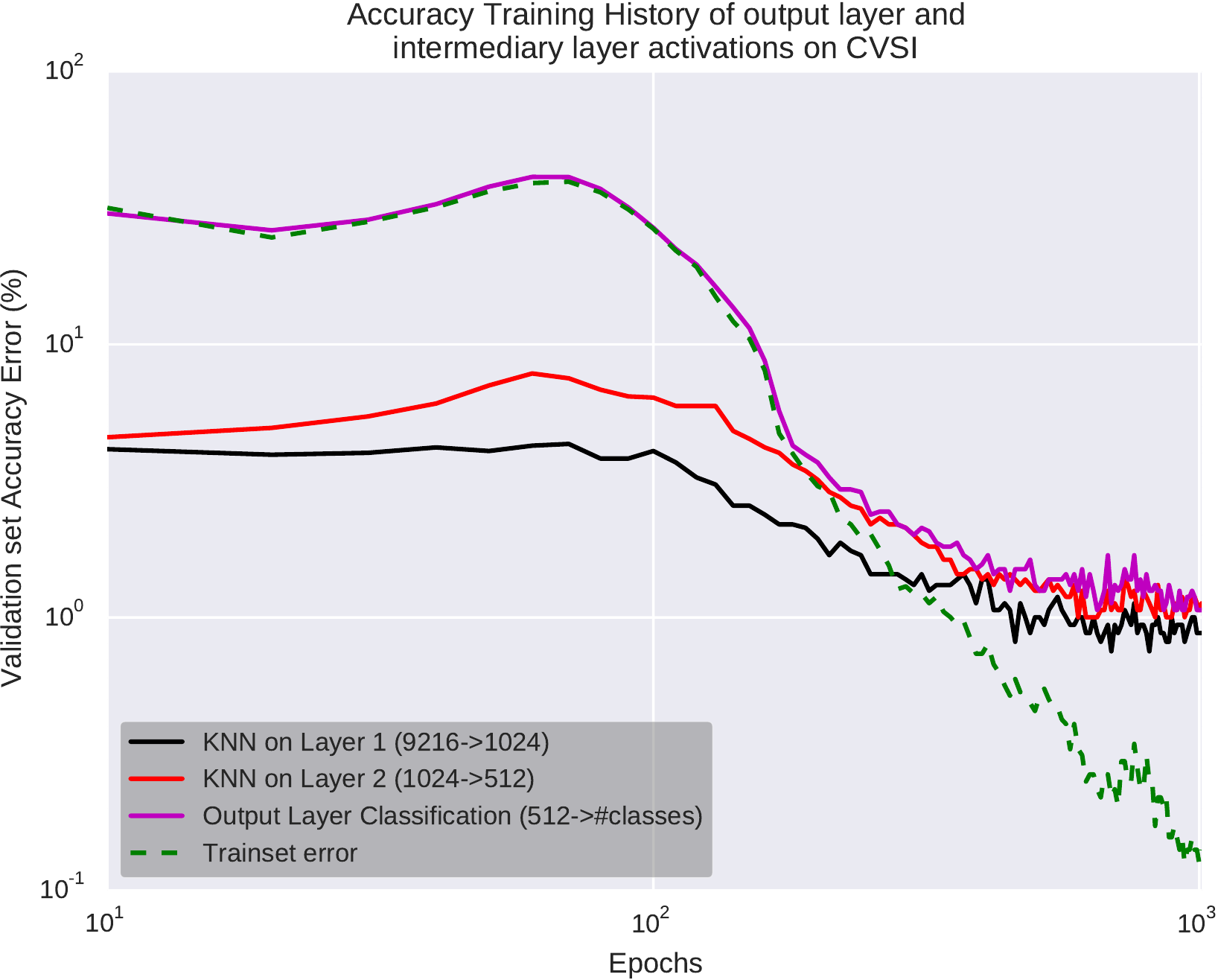}
\caption{Training of the MLP.}
\label{fig:history}
\end{figure}

While the discriminative deep MLP performs well, it is quite restricted by  the need of computational resources for training.
More than that, deep networks require datasets of substantial size and with all classes represented in a balanced way.
The other alternative is to use metric learning techniques, which can have some drawbacks.
Metric learning methods tend to have quadratic and even cubical complexities with respect to feature dimensionality, therefore an intermediary dimensionality reduction technique must also be used.
The idea of using neural networks dedicated to metric learning is best exemplified by the Siamese network architecture~\cite{bromley1993signature}.
While Siamese networks have all the benefits of metric learning in typical classification tasks, such as digit image classification, the results are lower than the state-of-the-art classifiers~\cite{liu2013probabilistic}.
On the other hand, intermediary activations of CNN are being used as generic feature extractors which are then classified with off the shelf classifiers such as Support Vector Machines (SVM)~\cite{razavian2014cnn}.
The work presented in this paper is greatly influenced by the principal idea in~\cite{razavian2014cnn} of using intermediary activations of deep CNN as generic features for standard classifiers in tasks other than what the original CNN was trained for.
Established CNN architectures do not directly preserve the aspect ratio of samples.
In the case of word samples, this means that the same letters could have a different representation if they appeared in words of different size.
There is no straight forward solution to this problem, i.e. the winning method of the CVSI competition addressed this problem by performing a sliding window of a fixed aspect ratio in each word and selecting the window with highest activation~\cite{cvsi2015}.
The drawback in such an approach is that all information outside the maximal activation window is ignored.
The authors propose the use of hand-crafted features that can address the aspect ratio problem.
Specifically the authors use the SRS-LBP histograms as inputs to a deep MLP since the pooling mechanism of the LBP histograms preserves perfectly the aspect ratio.
Building on the idea of~\cite{razavian2014cnn} the activations of the early layers in the MLP are used as input to a Nearest Neighbour classifier.
Depending on the dataset, lower levels of the proposed MLP can reach in performance and even exceed its output layer.
At the same time, networks used with this strategy are not limited to classes available during training.
In Fig.~\ref{fig:history} the error rates of all layers during training of the proposed MLP on the CVSI2015 dataset can be seen.

\section{Experiments}
\label{s:experiments}
In the experimental section we present experiments on script identification and language identification that demonstrate the potential of the proposed approach\footnote{Additional experimental resources are available at \url{http://nicolaou.homouniversalis.org/2016/01/07/visual_script.html}}.
\subsection{Video-text Script Identification}
\label{sec:video}
\begin{figure}
\begin{tabular}{ccc}
\includegraphics[width=.43\columnwidth]{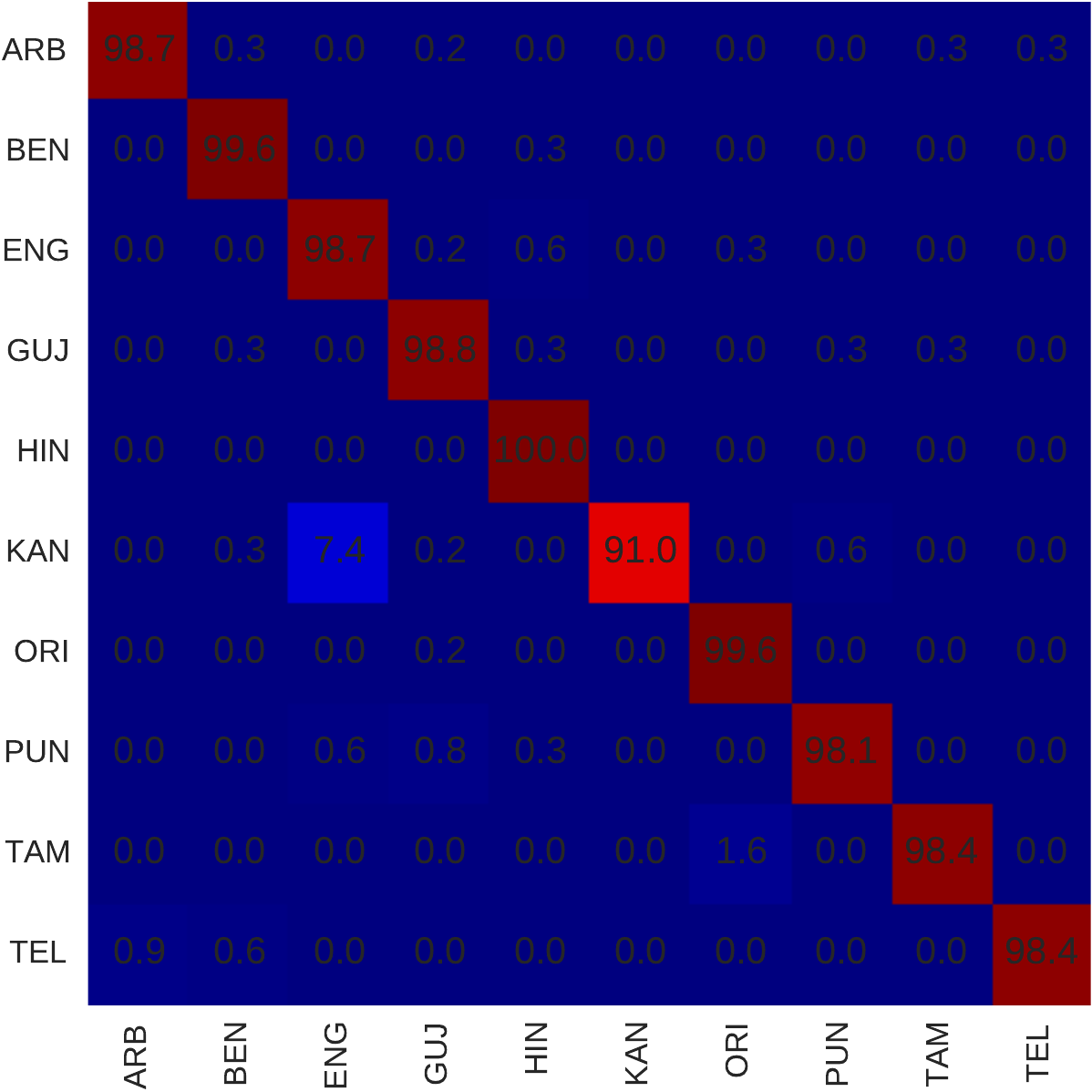} &
\includegraphics[width=.43\columnwidth]{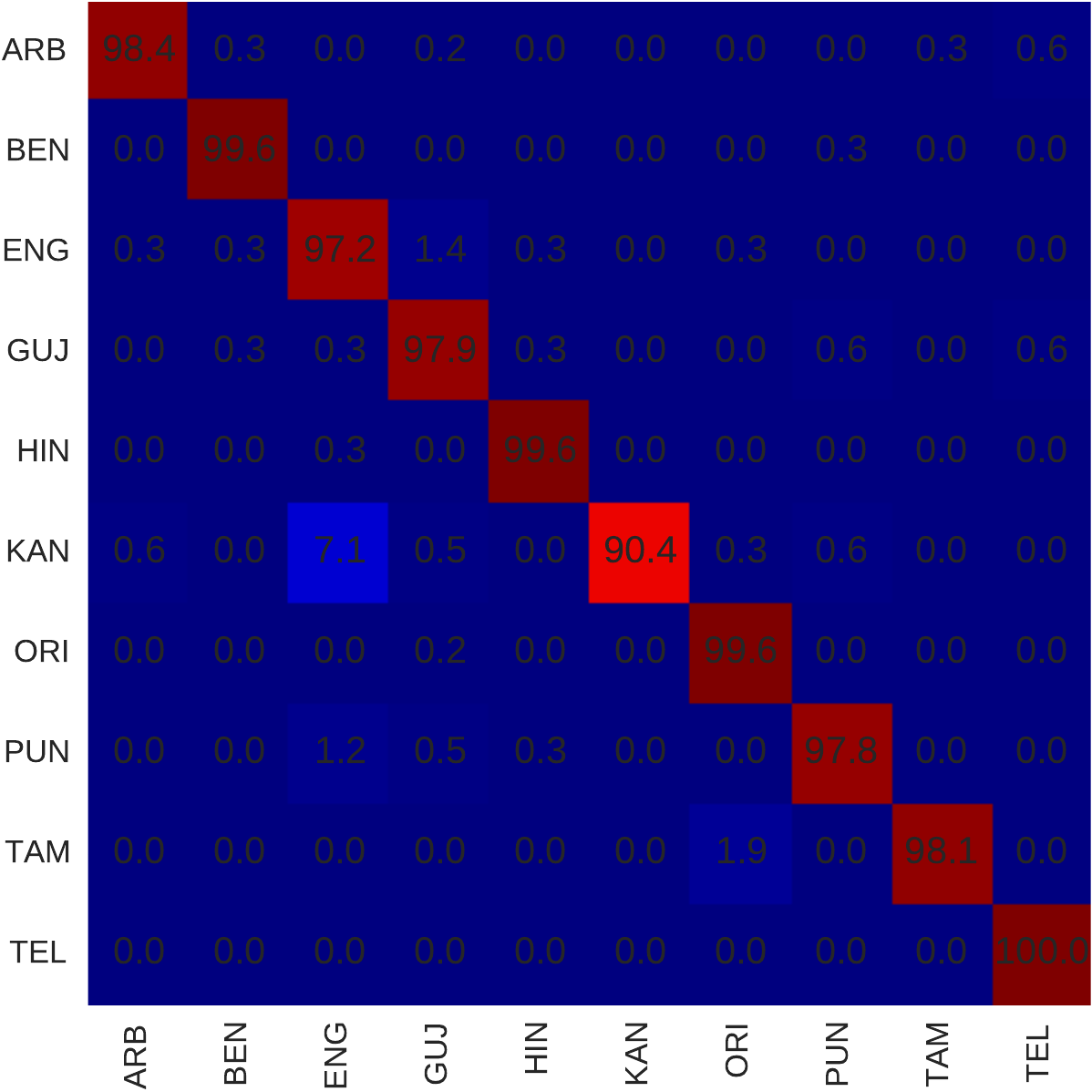}\\ (a) & (b) 
\end{tabular}
\caption{Confusion matrices for the CVSI dataset. Accuracy of the Nearest Neighbor for the activations of the first layer (a) and the third layer (b) }
\label{fig:confusion_cvsi}
\end{figure}

\input{tbl_cvsi}
The principal experiment to demonstrate near state-of-the-art performance is by comparing to the methods participating in the CVSI 2015 Video Script Identification~\cite{cvsi2015}.
The dataset contains of 10 languages used commonly in India: Arabic, Bengali, English, Gujrathi, Hindi, Kannada, Oriya, Punjabi, Tamil, Telugu.
The dataset consists of cropped images containing a single word each.
Most words appear to come from overlayed text, but there are also images that appear to be scene-text.
The dataset comes partitioned to a train-set, a test-set, a validation set, and a small sample-set.
For the experiments the test-set was isolated and used only for testing, the remaining data were mixed and partitioned randomly for training during the tuning of the proposed deep MLP architecture.
While the competition defines four tasks that are related to different use cases specific to India, such as discriminating between languages occurring on the same regions, in our experiments we only address Task-4, classifying all 10 scripts, as it is the most generic task.
In table~\ref{tbl:cvsi} the performance per script of every participant to the competition can be seen along with the accuracy achieved by each layer.
All layers of the proposed deep MLP rank on average second to the method submitted by Google.
While the method of Google, the state-of-the-art, obtains 98.9\% using a CNN, k-NN on the first layer of the deep MLP obtains 98.2\% while layer 2 obtains 97.3\% and the output layer obtains 97.9\%.
In Fig.~\ref{fig:confusion_cvsi} the confusion-matrices between languages for k-NN on the first layer, as well as the output layer can be seen. 
What stands out is the non-symmetric misclassification of 7\% English samples as Kannada; all other confusions could be considered negligible.
It can also be observed that layer 1 and layer 2  demonstrate some consistency.

\subsection{Scene-text Script Identification}
\label{s:scene}
While the method was developed for video-text script identification, experiments on how it would perform on script detection in the wild were performed.
We used the SIW dataset.
Two variants of the dataset are publicly available.
The SIW-10~\cite{shi2015automatic} contains cropped word images in Arabic, Chinese, English, Greek, Hebrew, Japanese, Korean, Russian, Thai, and Tibetan.
The SIW-13~\cite{shi2015script} adds Cambodian, Kannada, and Mongolian to the languages of SIW.
SIW-10 is partitioned in a train-set of 8,045 samples and test-set of 5,000 images while SIW-13 9,791 and 6,500 respectively.
Brief experimentation suggested that the partition of SIW-13 is not compatible with SIW-10, as test samples from SIW-13 appear to be in the SIW-10 train-set.
At the time of writing this paper the state-of-the art performance on the particular dataset is 94.6\% and is achieved by the MSPN method introduced in~\cite{shi2015automatic}.
Briefly, MSPN is a CNN developed specifically for script identification which introduces among other things a horizontal pooling layer.
\begin{figure}
\includegraphics[width=\columnwidth]{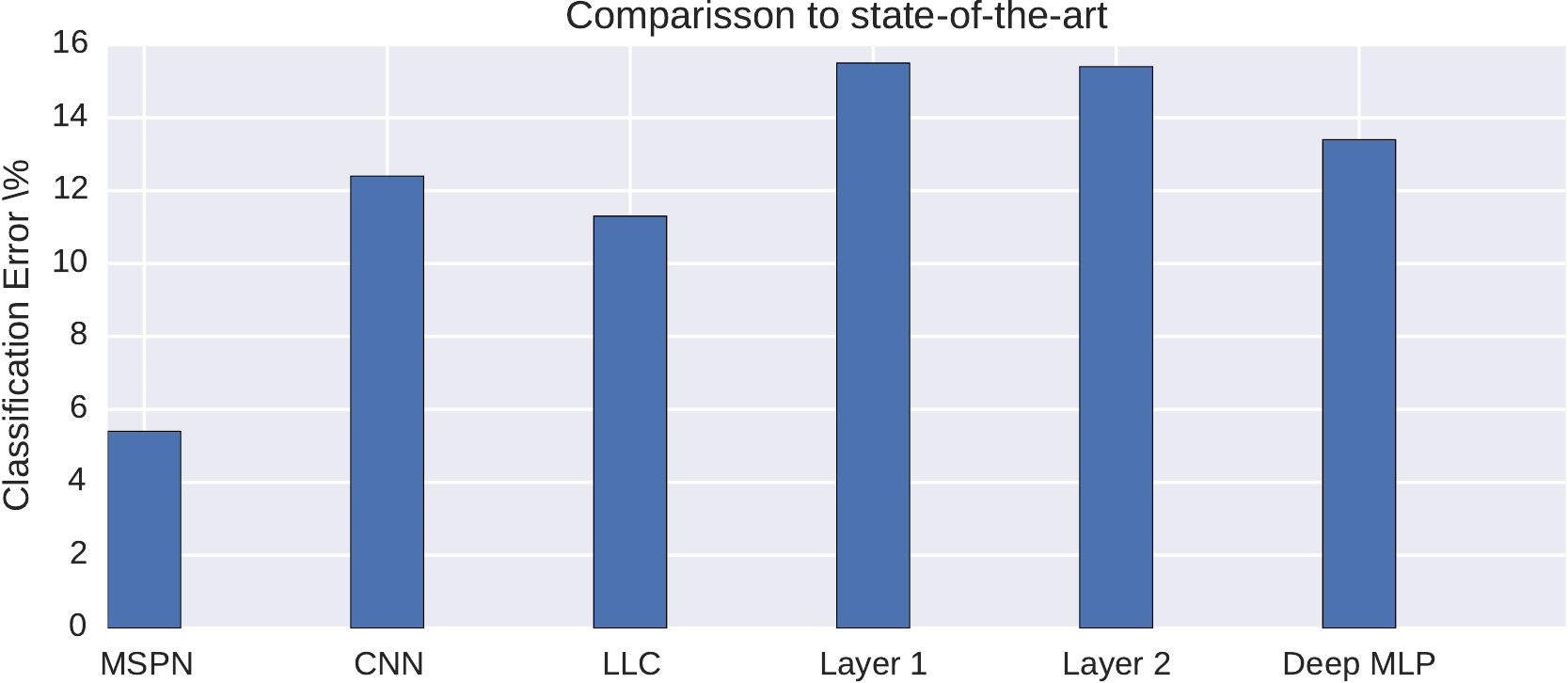}
\caption{Comparison to state-of-the-art on SIW-10. Error rates of the state-of-the-art CNN approach (MSPN), CNN baseline methods (CNN and LCC), intermediary layers as metric learning fed in to a Nearest Neighbour Classifier (Layer 1,Layer 2), and the proposed (Deep MLP).}
\label{fig:siw10_soa}
\end{figure}
In Fig.~\ref{fig:siw10_soa} a comparison of the proposed deep MLP with state-of-the-art methods for script detection in the wild can be seen.
The proposed method achieved an error rate of 13.4\% in classification accuracy which is significantly worst than the state-of-the-art 5.6\%.
Yet, this experiment allows an analysis in to the workings of the proposed deep MLP and the benefits of using k-NN on the intermediary activations.
The initial SIW-10 dataset was augmented by the three new languages of SIW-13.
An MLP trained on the SIW-10 was used to perform k-NN on the augmented dataset.

\begin{figure}
\includegraphics[width=\columnwidth]{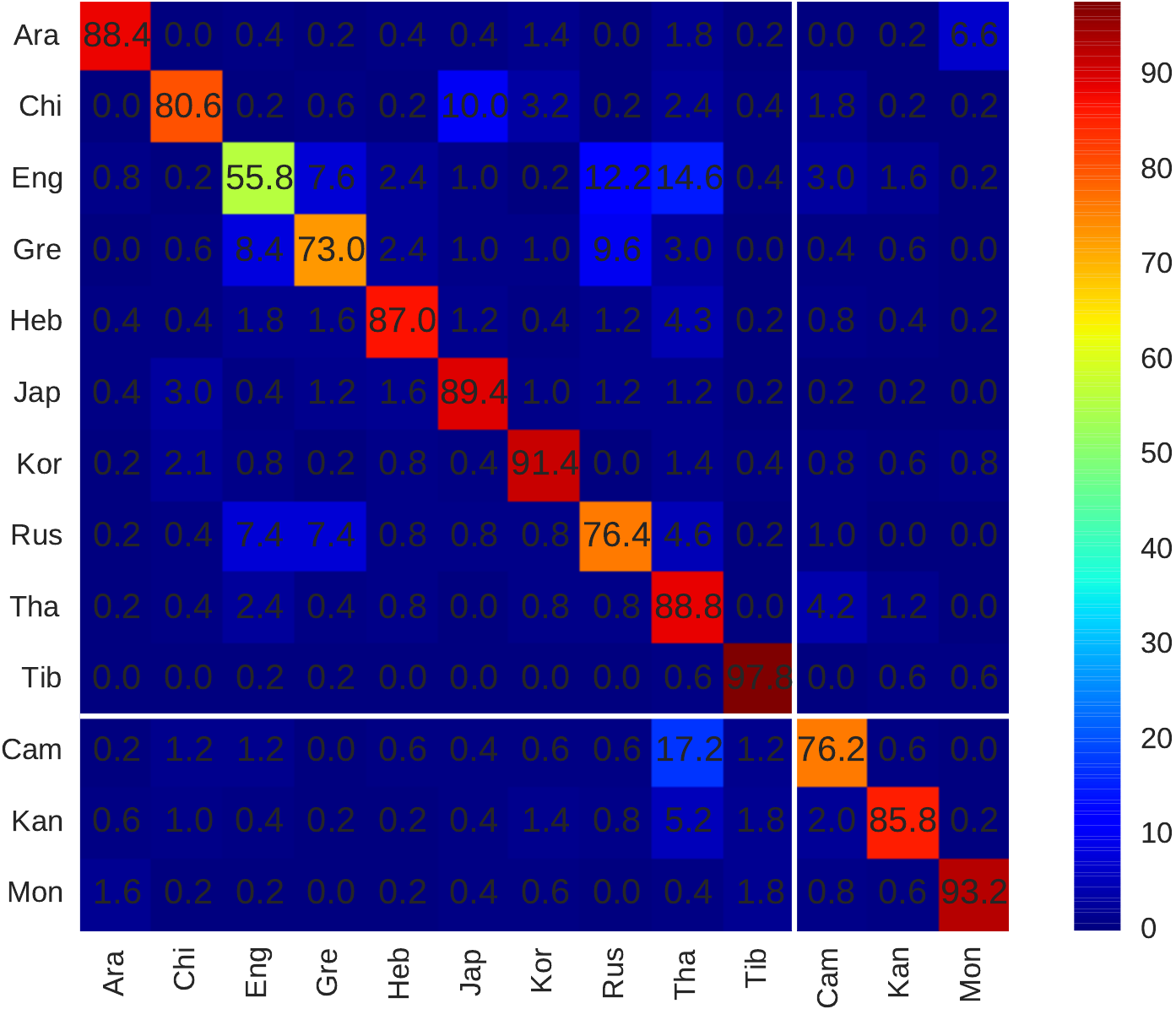}
\caption{Confusion matrix of k-NN on 13 scripts form the SIW datasets using the first layer of a deep MLP trained on 10 of them.}
\label{fig:siw10_confusion}
\end{figure}

In Fig.~\ref{fig:siw10_confusion} a confusion matrix of employing the first layer of the MLP with k-NN on the SIW-10 dataset augmented by the 3 languages of SIW-13.
The overall accuracy is 83.7\%, while for the initial 10 scripts it is 84.5\%.
While the second layer performed better than the first on the dataset for which the model was trained,  84.6\%, it proved to be less generic than the first layer and got 77.3\% when applied on all 13 classes.

The fact that the classes used to train the model have an average accuracy of 82.9\% and the unseen classes have an average accuracy of 85.1\% demonstrates the overall genericness of the first layer.
As opposed to the CVSI experiments, when training on SIW data, consistently the second layer seemed to outperform the first layer after some epochs.

\begin{figure}
\includegraphics[width=\columnwidth]{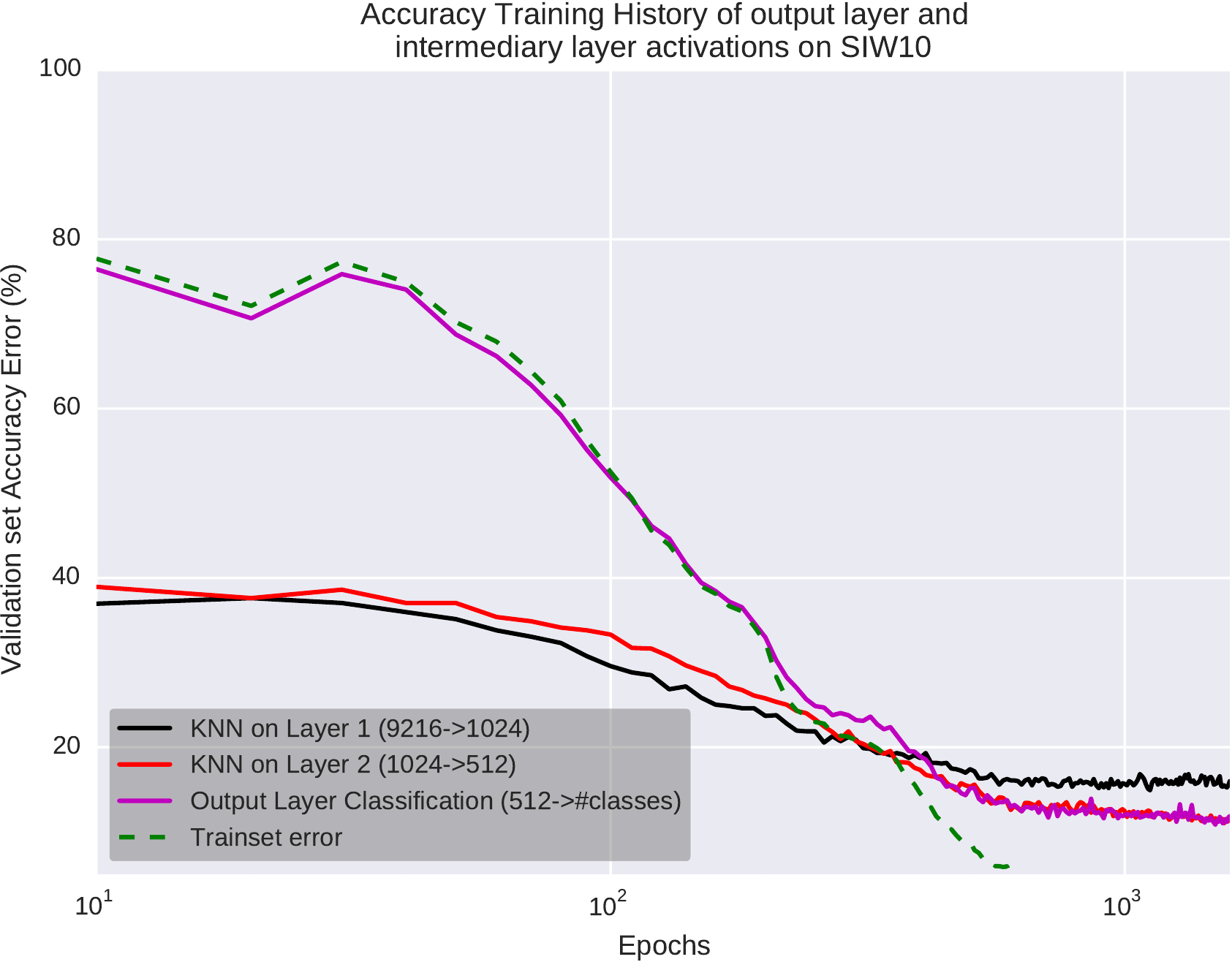}
\caption{Training of the proposed deep MLP on the SIW-10 data.}
\label{fig:siw10_history}
\end{figure}

In Fig.~\ref{fig:siw10_history} the training of the deep MLP can be seen and we can observe that the second layer converges towards the output layer while the first layer appears to be more independent.
The extent to which layer 2 is domain specific while layer 1 is much more domain independent can be seen in table~\ref{tbl:crossdomain}, where layer 1 increases error rates when changing domains by 2.2 and 2.8 times, while layer 2 increases error rates by 3.4 and 8.9 times respectively.
\begin{table}
\centering
\caption{Crosss-domain use of deep MLP layers}
\begin{tabular}{cccc}
\hline
MLP Train & Retrieval & k-NN on & k-NN on \\
Dataset &  Dataset &  layer 1 &  layer 2 \\ \hline
SIW-10 & CVSI & 94.8\% & 76.1\% \\
CVSI & CVSI & 98.2\%  & 97.3\% \\ \hline

CVSI & SIW-10 & 66.4\% & 47.7\% \\
SIW-10 & SIW-10 & 84.5\% & 84.6\% \\\hline
\end{tabular}
\label{tbl:crossdomain}
\end{table}

\subsection{Visual Identification of Handwritten Language}
\label{s:visual}
\begin{figure*}
\centering
\begin{tabular}{cccc}
\includegraphics[width=.2\textwidth]{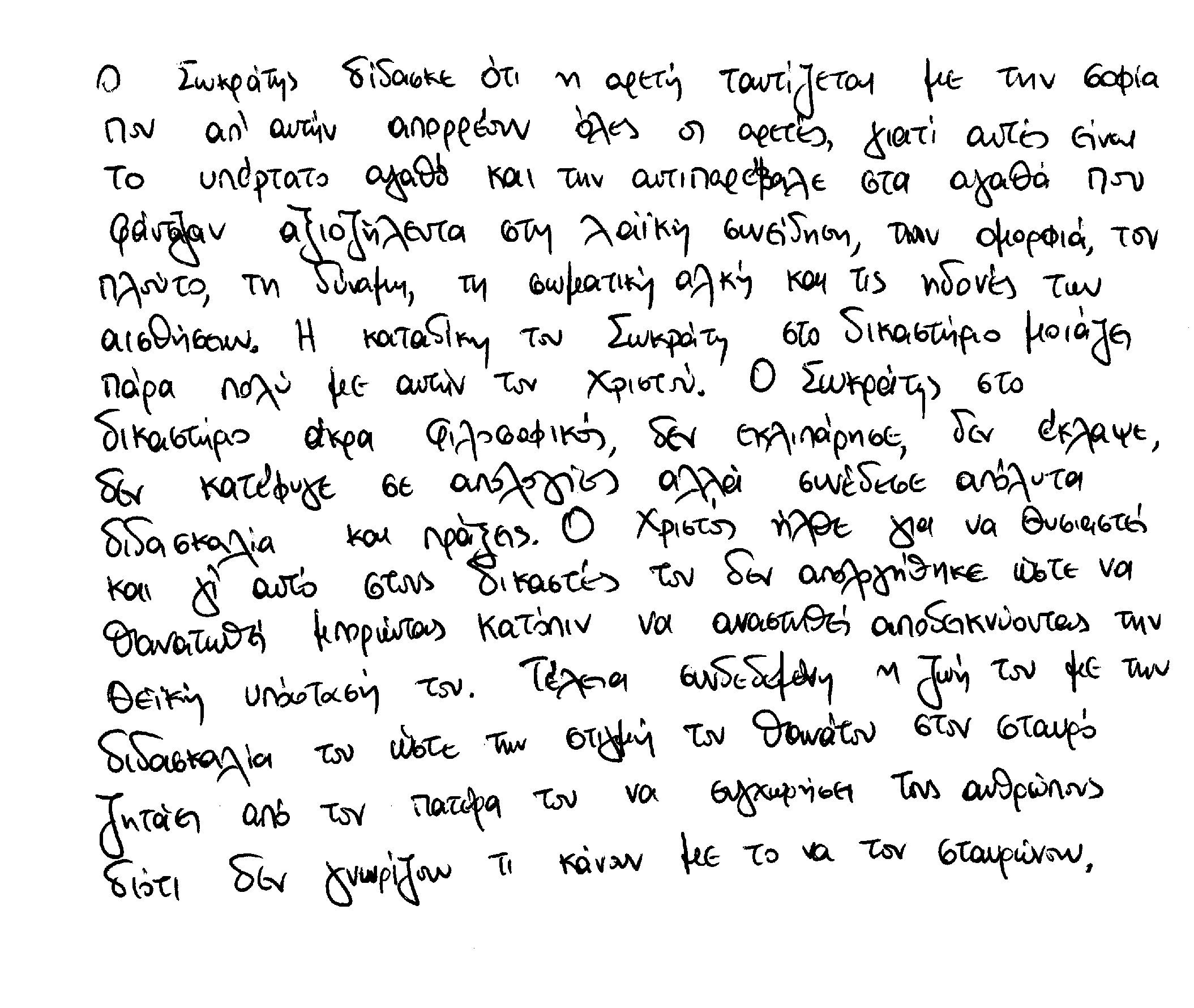} &
\includegraphics[width=.2\textwidth]{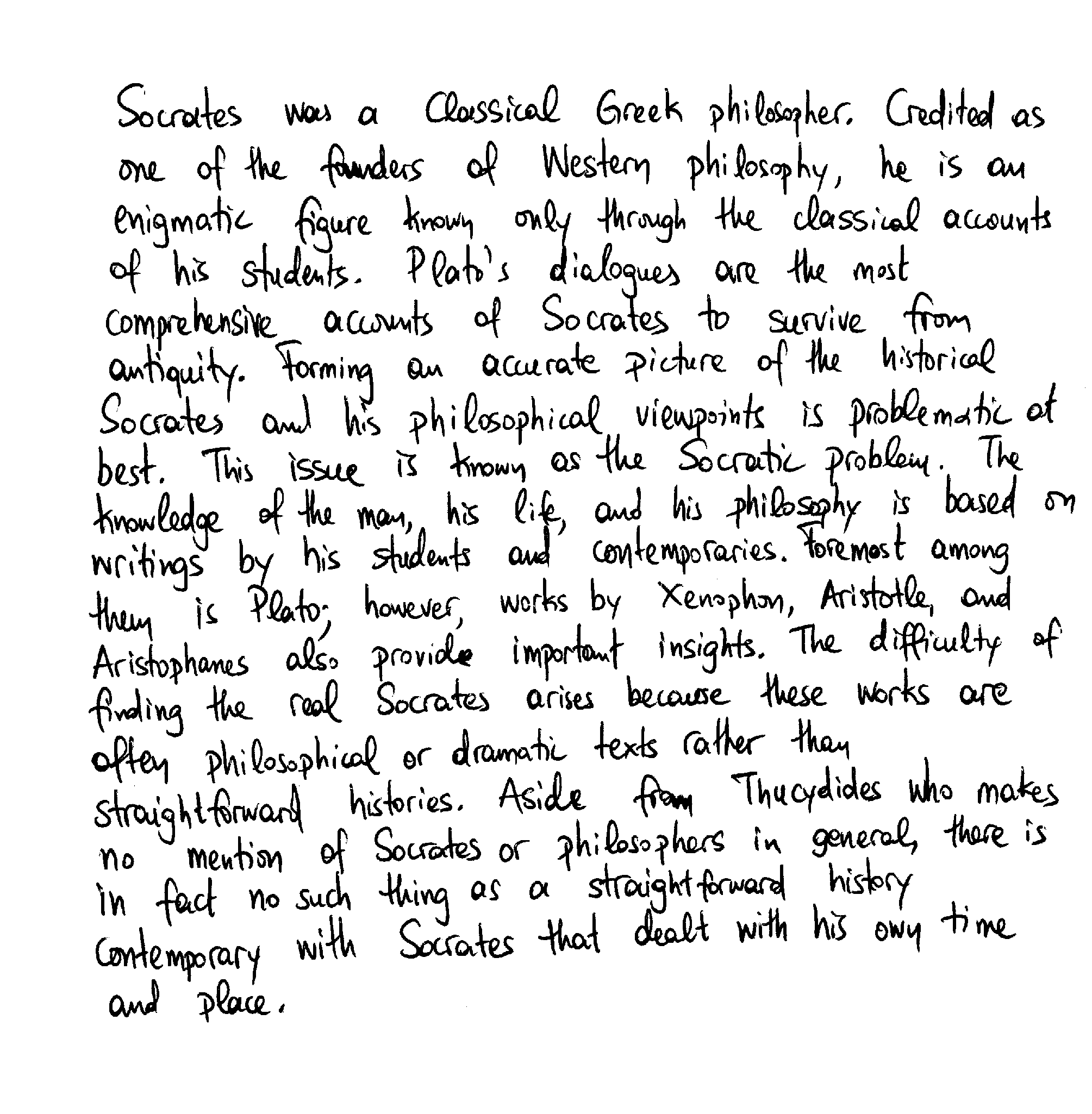} &
\includegraphics[width=.2\textwidth]{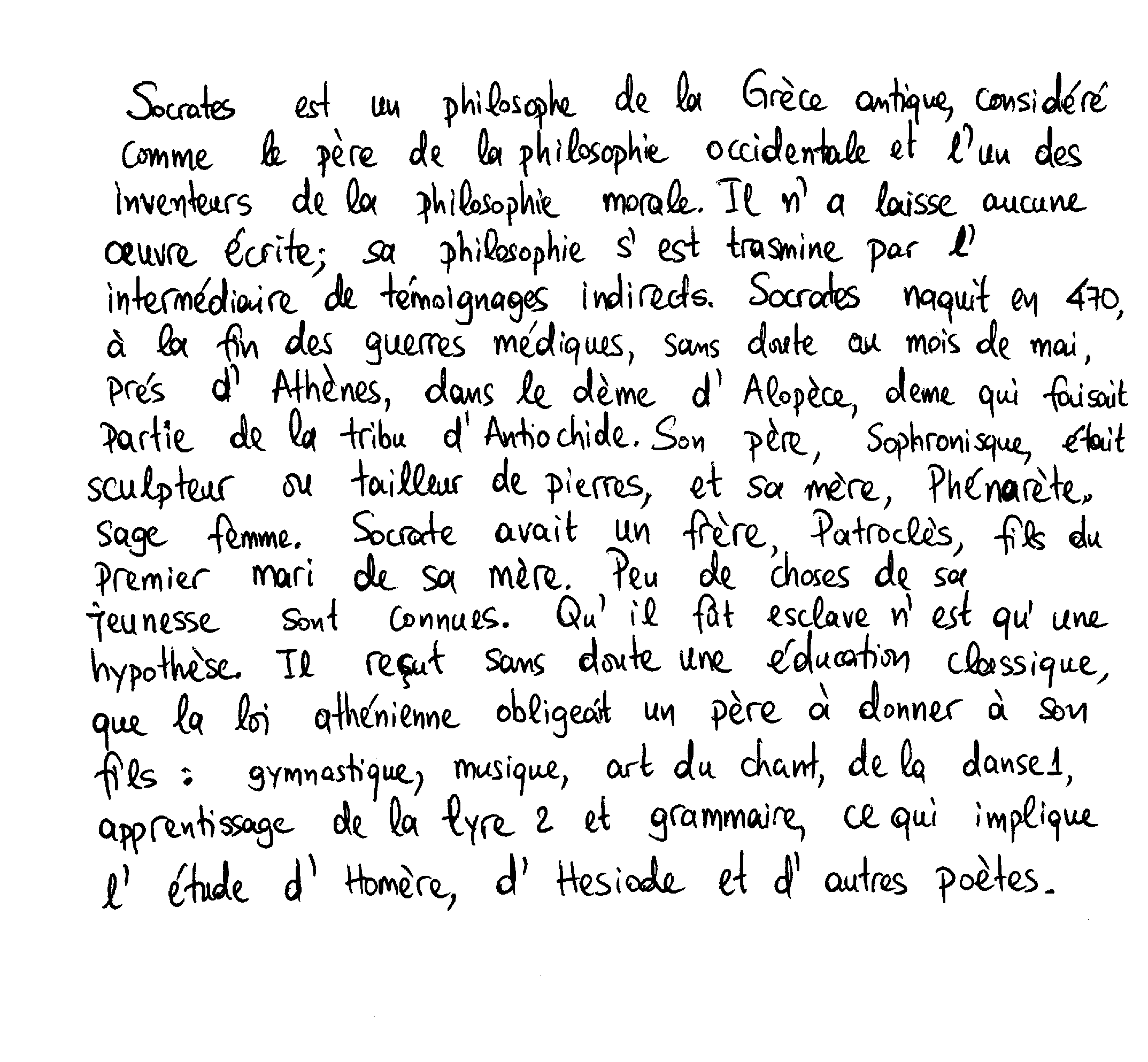} &
\includegraphics[width=.2\textwidth]{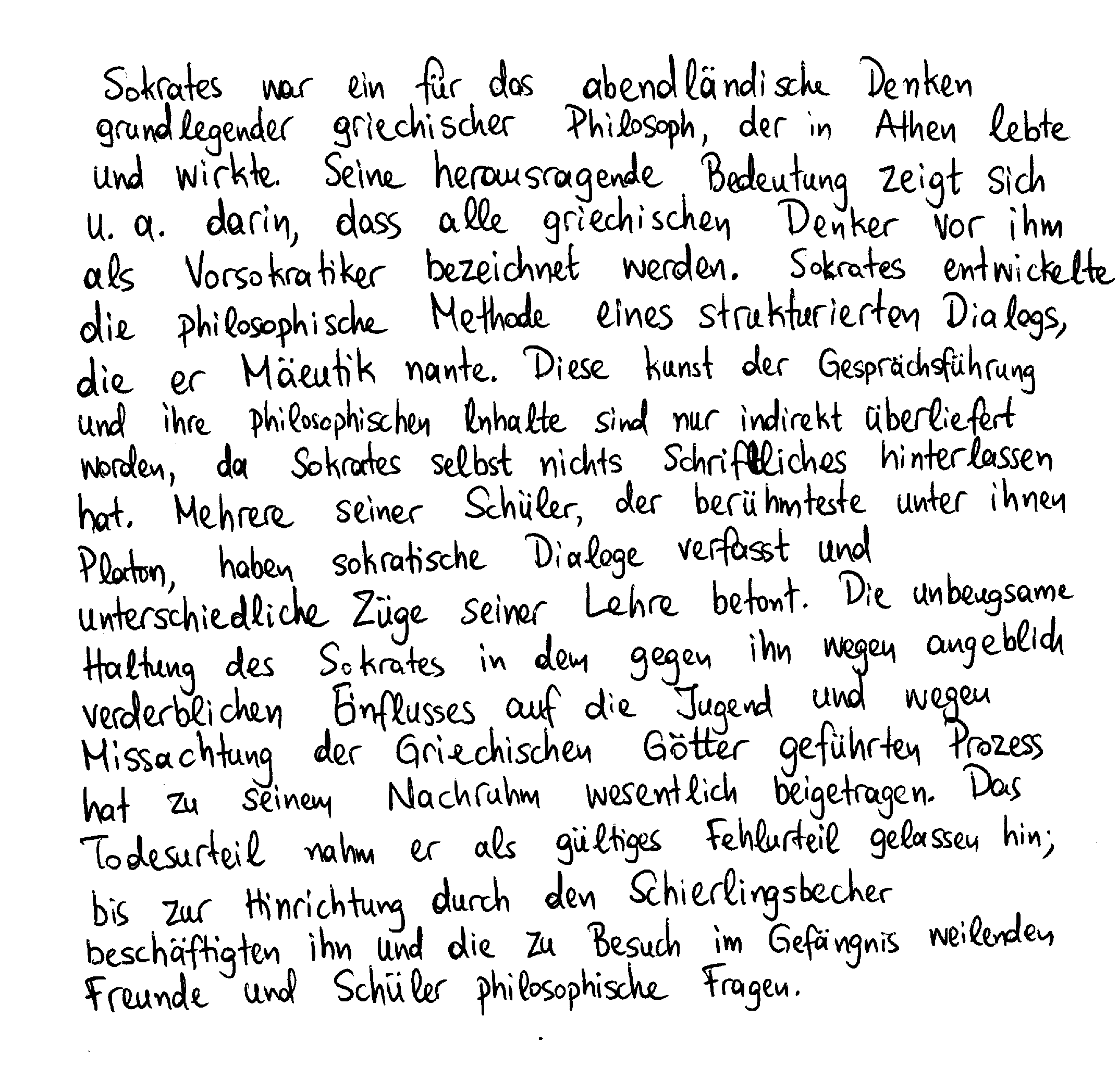} 
\\ (a) & (b) & (c) & (d)
\end{tabular}
\caption{Samples from handwritten language identification. Samples of the same text and writer in Greek (a), English (b), French (c), and German (d).}
\label{fig:wi_samples}
\end{figure*}

The boundary between script and language identification is hard to define, as can be best exemplified in Latin derived languages.
Visual language identification could also be perceived as fine-grained script classification.
Yet distinguishing between such scripts before identification is required if one is to use language models for identification.
In the case of handwriting identification, it becomes even more important, since identification frequently relies in word-spotting, which by definition needs a lexicon.
In order to address this problem the ICDAR 2011 writer identification dataset was used\cite{icdar2011} to estimate the language classification.
This dataset consists of two paragraph-long texts translated to four languages: Greek, English, French, and German.
In Fig.~\ref{fig:wi_samples} the same text written by the same writer in all four languages can be seen.
Twenty six writers wrote all these samples which were then digitized and binarized.
State-of-the-art methods report performances of over 95\% accuracy in writer identification.
As the dataset has never been used in the language identification context and visual handwritten-text language identification is a new problem to the authors knowledge, there is no state-of-the-art method.
In order to make writer identification irrelevant, a 26-fold cross validation scheme was employed.
All 8 samples contributed from every writer were used as testing samples, while all other samples were used for training.
 
\begin{table}
\centering
\caption{Visual Language Identification Accuracy}
\begin{tabular}{l|c}
\hline
Method & Accuracy \\ \hline
Random Classifier & 25.0\% \\
SRS-LBP learning free pipeline & 50.96\% \\
SVM + SRS-LBP features & 91.18\% \\ 
Deep MLP + SRS-LBP features & \textbf{92.78}\% \\ \hline
\end{tabular}
\label{tbl:language}
\end{table}

\begin{figure}
\centering
\includegraphics[width=.6\columnwidth]{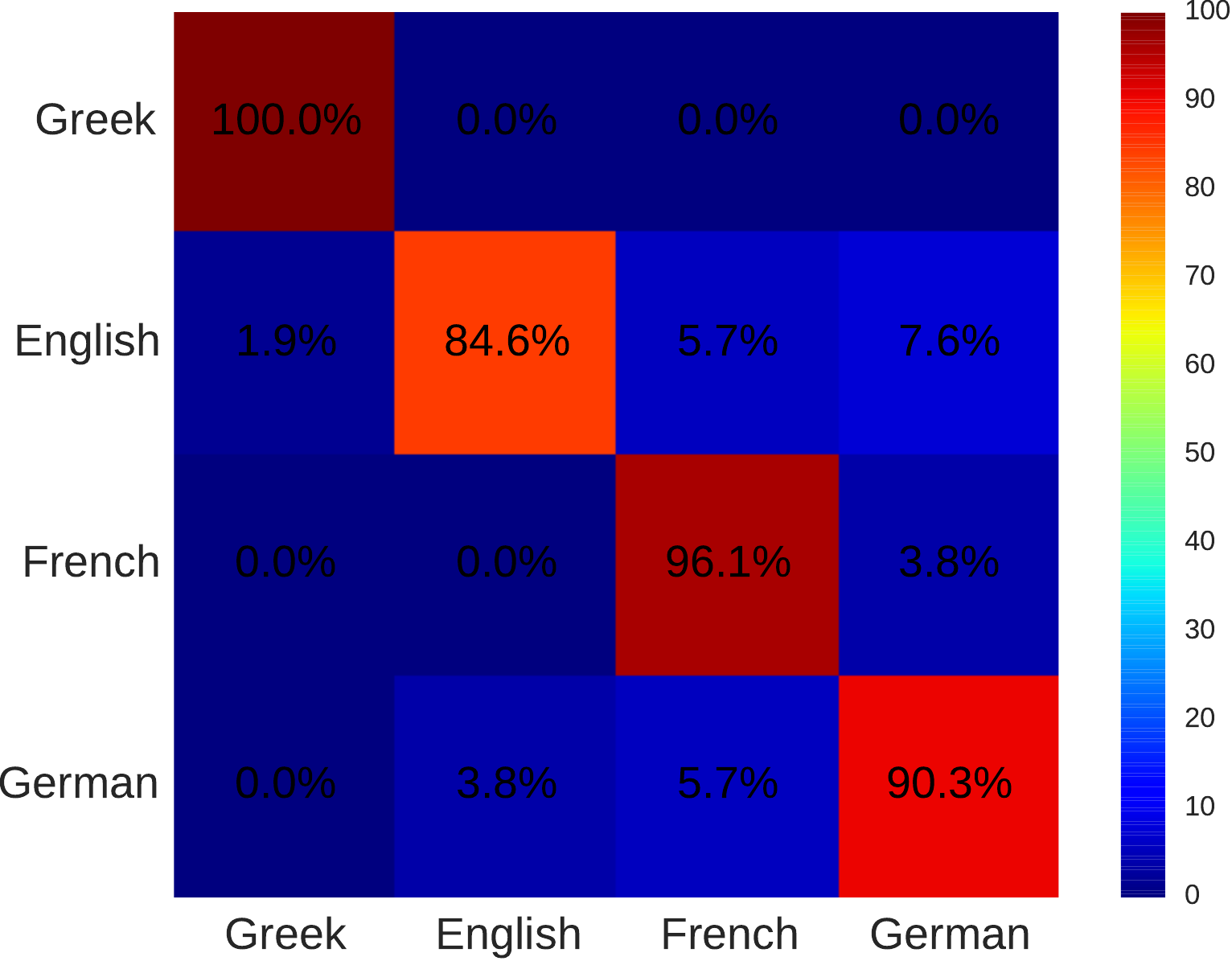}
\caption{Confusion Matrix on Visual Language Detection with an ANN 26-fold cross-validation}
\label{fig:confusion_wi_ann.pdf}
\end{figure}
In table~\ref{tbl:language} the performance of the proposed deep MLP along with baselines is presented.
The dataset is totally balanced and has 4 classes, so an unbiased random classifier would be performing with 25\%.
The SRS-LBP unsupervised learning from~\cite{nicolaou2015sparse} performs poorly, although significantly better than the random classifier.
The same pipeline when applied on the same dataset for writer identification obtains 98.1\%.
This could be interpreted as an indication of how harder the Handwritten Visual Language Identification problem is compared to writer identification, at least for LBP features.
The proposed method Deep MLP is exactly the same as the one described and used for CVSI, but instead of three pooling zones only global pooling is employed as the image has more than one text-line.
Deep MLP achieves top performance 92.78\%, although an SVM applied on the same features performs nearly as well.
In Fig.~\ref{fig:confusion_wi_ann.pdf} the confusion matrix of visual language classification can be seen.
As one would expect, Greek is separated from the other three perfectly while English, French, and German have some confusions.
It should be pointed out the dataset was acquired in Greece and all subjects would have Greek as their primary language and this might be helping distinguish it from the other three languages.
\section{Conclusions}
\label{s:conclusions}

\subsection{Discussion and Remarks}
Several conclusions can be drawn from the experiments.

As the problem exemplified by CVSI was the primary focus for the development of the proposed method, the  near state-of-the-art results validate the over-all strategy of using hand crafted texture features as the basis for script classification.
Although all top achieving methods in CVSI used CNN, the proposed method demonstrates that hand crafted features can outperform them.
It could be suggested that the fixed aspect ratio CNN requires, although addressed by all CNN methods in different ways, it is not as effective as the pooling performed on the LBP features.
Video-text as a phenomenon is a lot more regular, occlusions and other complicated noise and distortions are very rare for such data; it stands to reason that the SRS-LBP features perform better in this case compared to text in the wild.

Although there is a lack of competing methods on the task, the visual language identification experiments demonstrated that detecting the language before recognition is possible.
While the experiment was quite challenging, the fact that the dataset was balanced in every aspect might mean that the reported performance might not be the same in real world scenarios.
The fact that all texts are of the same size, acquired under exactly the same conditions etc., allows for a model so big to be trained on a mere 208 samples.
The model could be trained on more realistic data but then a much larger dataset would be required.
The high performance demonstrated by the proposed pipeline should probably be attributed to the SRS-LBP features rather than the deep MLP as an SVM classifier on the same features provided results nearly as high. 

When training the deep MLP on CVSI and SIW datasets, on each separate dataset the networks demonstrated consistency within the dataset.
On SIW data, after some epochs the output layer outperformed k-NN on layer~2 which in turn outperformed k-NN on layer~1.
On CVSI data, the behaviour was consistent.
Applying k-NN on layer~1 outperformed the output layer which in turn outperformed k-NN on layer~2.
The consistency in the difference between these two datasets suggests that video-text and scene-text script identification are different on how complicated they are and a deeper architecture might improve performance on scene-text.
The use of early layers in deep MLP as a learned metric is probably the most interesting contribution in this paper.
It should be noted, that during several training of the models in order to tune the architecture, the convergence of the curves was consistent for each dataset.
This consistency could be attributed to the drop-out regularization.
Drop-out regularization should also be credited with the fact that no over-fitting occurred on the trained data, since continuing training never increased the validation error in any significant manner.
The experiment on the augmented SIW-10 dataset provided an experimental and quantitative validation of a hypothesis underlying presented work:
In a sequence of fully connected layers, activations of earlier stages are more generic than later ones.
\label{s:discussion}
\subsection{Future Work}
\label{s:future}
This paper presents a work in progress and several questions remain open.
Further experimentation should be done in order to get a better understanding of the benefits and limits this metric learning has when compared to other methods such as Siamese networks.
Although the performance in script identification in the wild was high, it was significantly lower than the state-of-the-art.
Adapting a network architecture that performs better on unconstrained scene-text is one of the ways to follow on the reported work.
\newcommand{\BIBdecl}{\setlength{\itemsep}{0.10 em}}
\bibliographystyle{IEEEtran}
\bibliography{script}
\end{document}

%% file: fig_features.tex
\begin{figure*}
\centering
\begin{tabular}{lcccc}
 &(a) &  (b) & (c) & (d) \\
1)&
\includegraphics[align=c,width=4cm]{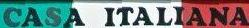} &
\includegraphics[align=c,width=4cm]{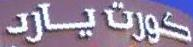} &
\includegraphics[align=c,width=4cm]{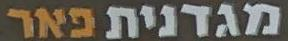} &
\includegraphics[align=c,width=4cm]{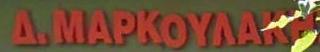}\\
2)&
\includegraphics[align=c,width=4cm]{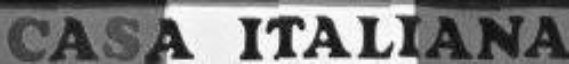} &
\includegraphics[align=c,width=4cm]{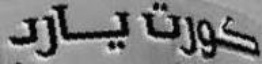} &
\includegraphics[align=c,width=4cm]{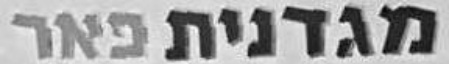} &
\includegraphics[align=c,width=4cm]{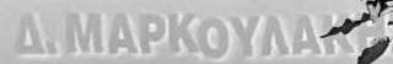}\\
3)&
\includegraphics[align=c,width=4cm]{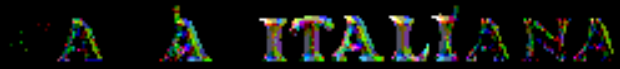} &
\includegraphics[align=c,width=4cm]{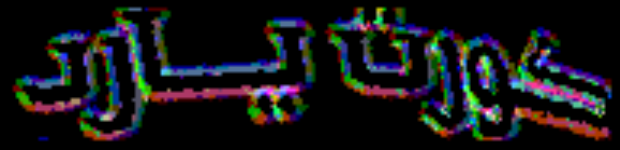} &
\includegraphics[align=c,width=4cm]{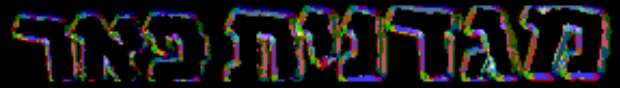} &
\includegraphics[align=c,width=4cm]{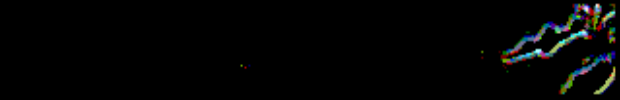}\\
4)&
\includegraphics[align=c,width=4cm]{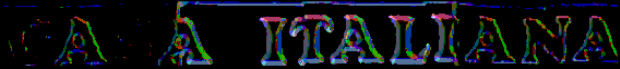} &
\includegraphics[align=c,width=4cm]{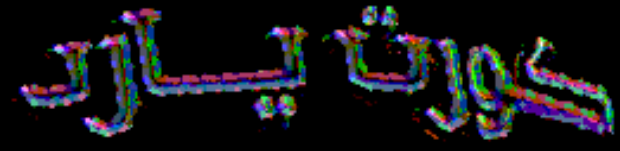} &
\includegraphics[align=c,width=4cm]{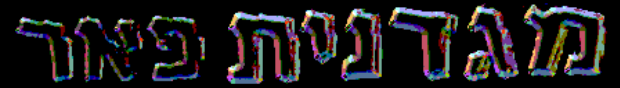} &
\includegraphics[align=c,width=4cm]{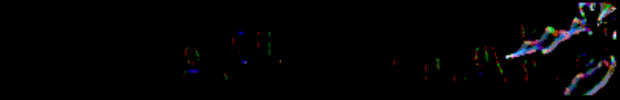}\\
\end{tabular}
\caption{Preprocessing and SRS-LBP transform for radius of 3. Data taken from SIW-10 dataset \cite{shi2015automatic}. 1) Original image, 2)Preprocessed image, 3)$\text{SRS-LBP}_{8,3}$ of input images and 4)$\text{SRS-LBP}_{8,3}$ of preprocessed images. }
\label{fig:features}
\end{figure*}

%% file: tbl_cvsi.tex
\begin{table*}
\centering
\caption{Accuracy \% on the CVSI Video-text Dataset}
\label{tbl:cvsi}
\begin{tabular}{l|ccccc|ccc}
\hline
Language & C-DAC & HUST & CVC & Google & CUK &  Layer 1, 1NN & Layer 2, 1NN & Layer 3 \\
\hline 
Arabic & 97.69 & \textbf{100.0} & 99.67 & \textbf{100.0} & 89.44 & 98.7 & 98.4 & 98.4\\
Bengali & 91.61 & 95.81 & 92.58 & 99.35 & 68.71 & \textbf{99.6} & 99.3 & \textbf{99.6}\\
English & 68.33 & 93.55 & 88.86 & 97.95 & 65.69 & \textbf{98.7} & 98.4 & 97.2\\
Gujrathi & 88.99 & 97.55 & 98.17 & 98.17 & 73.39 & \textbf{98.8} & 95.3 & 97.2\\
Hindi & 71.47 & 96.31 & 96.01 & 99.08 & 61.66 & \textbf{100.0} & 99.6 & 99.6\\
Kannada & 68.47 & 92.68 & 97.13 & \textbf{97.77} & 71.66 & 91.0 & 87.5 & 90.4\\
Oriya & 88.04 & 98.47 & 98.16 & 98.47 & 79.14 & \textbf{99.6} & \textbf{99.6} & \textbf{99.6}\\
Punjabi & 90.51 & 97.15 & 96.52 & \textbf{99.38} & 82.55 & 98.1 & 97.8 & 97.8\\
Tamil & 91.90 & 97.82 & \textbf{99.69} & 99.37 & 82.55 & 98.4 & 98.1 & 98.1\\
Telugu & 91.33 & 97.83 & 93.80 & 99.69 & 57.89 & 98.4 & 98.1 & \textbf{100.0}\\ \hline
Average & 84.66 & 96.69 & 96.00 & \textbf{98.91} & 74.06 & 98.18 & 97.26 & 97.9\\ 
\hline
\end{tabular}
\end{table*}